\PassOptionsToPackage{hyphens}{url}
\PassOptionsToPackage{most}{tcolorbox}
\documentclass[11pt]{article}
\let\standardaddcontentsline\addcontentsline
\AddToHook{package/hyperref/before}{\let\addcontentsline\standardaddcontentsline}
\usepackage[preprint]{arxiv}
\usepackage{times}
\usepackage[utf8]{inputenc}
\usepackage[T1]{fontenc}
\usepackage{microtype}
\usepackage{placeins}
\usepackage{amsmath,amssymb,bm}
\usepackage{graphicx,booktabs,multirow}

\usepackage{algorithm}
\usepackage{algpseudocode}

\newcommand{\tablehead}[1]{\textbf{\boldmath#1}}
\newcommand{\papertablesetup}{\fontsize{9}{10.5}\selectfont
  \setlength{\tabcolsep}{3pt}\renewcommand{\arraystretch}{1}\setlength{\heavyrulewidth}{0.6pt}\setlength{\lightrulewidth}{0.4pt}\setlength{\cmidrulewidth}{0.3pt}\setlength{\aboverulesep}{2pt}\setlength{\belowrulesep}{2pt}}
\usepackage{tikz}
\usetikzlibrary{arrows.meta,positioning,calc,fit}
\definecolor{draftred}{RGB}{155,52,45}
\definecolor{graphgreen}{RGB}{43,113,91}
\definecolor{graphmemory}{RGB}{124,91,164}
\definecolor{graphtarget}{RGB}{205,139,57}
\definecolor{resultbest}{RGB}{244,205,202}
\definecolor{resultsecond}{RGB}{199,216,238}
\newcommand{\resultbox}[2]{\begingroup\setlength{\fboxsep}{0.75pt}\raisebox{0pt}[0pt][0pt]{\colorbox{#1}{#2}}\endgroup}
\newcommand{\resultbest}[1]{\resultbox{resultbest}{\textbf{#1}}}
\newcommand{\resultsecond}[1]{\resultbox{resultsecond}{\underline{#1}}}

\newcommand{\Attn}{\operatorname{Attn}}

\newcommand{\cat}{\mathbin{\Vert}}

\hypersetup{
  hypertexnames=false,
  pdftitle={WeaveVLA: 3D Relational Grounding for Vision-Language-Action Models},
  pdfauthor={Zien Wang, Yuhao Pan, Yunfeng Fan, Nan Cheng, Xin Xia, Wenchao Xu}
}
\newcommand{\correspondingemail}{wenchaoxu@ust.hk}
\title{WeaveVLA: 3D Relational Grounding for Vision-Language-Action Models}
\author{\normalfont
  Zien Wang\textsuperscript{1,*}\quad
  Yuhao Pan\textsuperscript{1,*}\quad
  Yunfeng Fan\textsuperscript{2}\quad
  Nan Cheng\textsuperscript{3}\quad
  Xin Xia\textsuperscript{4}\quad
  Wenchao Xu\textsuperscript{1,\ensuremath{\dagger}}\\[3pt]
  \normalfont\small
  \textsuperscript{1}The Hong Kong University of Science and Technology\quad
  \textsuperscript{2}Hong Kong Polytechnic University\\
  \normalfont\small
  \textsuperscript{3}Xidian University\quad
  \textsuperscript{4}Zhejiang University\\[2pt]
  \normalfont\small\textsuperscript{*}\,Equal contribution.\quad
  \textsuperscript{\ensuremath{\dagger}}\,Corresponding author.
}
\begin{document}
\setlength{\headheight}{24pt}
\fancypagestyle{firstpage}{
  \fancyhf{}
  \rhead{\small\textcolor[HTML]{4D4D4D}{\today\quad Preprint}}
  \cfoot{\thepage}
  \renewcommand{\headrulewidth}{0.4pt}
}
\maketitle
\vspace{-5mm}

\begin{abstract}

Vision-language-action (VLA) models leverage pretrained vision-language representations for instruction-conditioned robot control, yet their understanding of task-relevant 3D relations and how these relations evolve during manipulation remains largely implicit.
To address this challenge, we introduce \textbf{WeaveVLA}, a framework that enhances VLA action generation by explicitly modeling task-relevant relative 3D relations and maintaining relational context across observations.
WeaveVLA employs a task-conditioned sparse graph neural network to encode relative 3D relations among objects, the gripper, and nearby references.
Pair-specific recurrent edge states integrate geometric evidence across observations to maintain evolving spatial context.
A learned-query readout summarizes the updated graph representations into compact relational tokens, which condition the action expert alongside the original vision-language context. An auxiliary geometry reconstruction objective further encourages spatial structure preservation.
We instantiate WeaveVLA on StarVLA-GR00T and $\pi_{0.5}$, achieving over 98\% average success on LIBERO with both backbones. On LIBERO-Plus, WeaveVLA improves success from 74.9\% to 82.0\% and from 85.7\% to 87.8\%, respectively, with additional improvements on CALVIN D$\rightarrow$D.
On three real-world tasks, the StarVLA-GR00T-based variant improves average success from 55.6\% to 70.0\% under standard conditions and from 33.3\% to 53.3\% under OOD perturbations, surpassing the $\pi_{0.5}$ baseline in both settings.

\par\medskip
\noindent{\small
\begin{tabular}{@{}l@{\qquad}l@{}}
\textbf{Correspondence:} & \href{mailto:\correspondingemail}{\texttt{\correspondingemail}} \\
\textbf{Project Page:} & \href{https://kingzein.github.io/weave-vla/#overview}{\texttt{kingzein.github.io/weave-vla}}
\end{tabular}
}

\end{abstract}

\addtocontents{toc}{\protect\setcounter{tocdepth}{-1}}

\begin{figure}[!t]
\centering
\includegraphics[width=\linewidth]{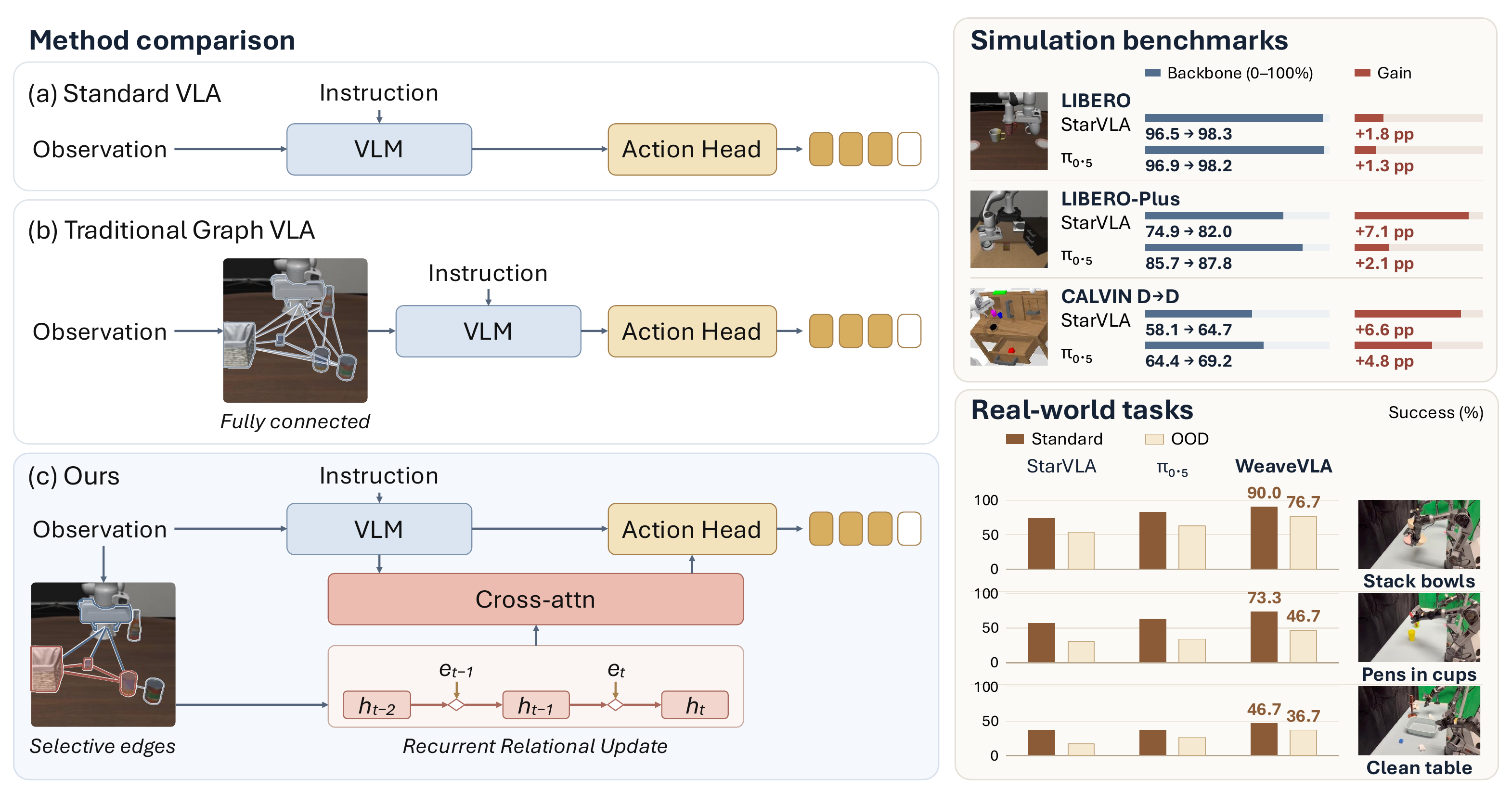}
\caption{Conceptual comparison of (a) a standard VLA, (b) an illustrative fully connected graph-based VLA, and (c) WeaveVLA. Unlike uniform scene-wide connectivity, WeaveVLA selects task-relevant connections and updates pair-specific relational states across observations.}
\label{fig:teaser}
\end{figure}

\section{Introduction}
\label{sec:intro}

Vision-language-action (VLA) models combine pretrained vision-language representations with robot demonstrations to learn instruction-conditioned manipulation policies.
Representative approaches such as OpenVLA and OpenVLA-OFT have advanced general-purpose manipulation, while recent systems have expanded these capabilities across robotic embodiments~\cite{kim2024openvla,kim2025fine,nvidia2026gr00tn17}.
Flow-based policies, including $\pi_0$, $\pi_{0.5}$, and $\pi_{0.7}$, further advance continuous action generation, generalization, and steerable control~\cite{black2024pi0,intelligence2025pi_,intelligence2026pi07steerablegeneralistrobotic}.
Despite these advances, understanding and maintaining the spatial relationships that govern manipulation remains an important challenge.

Robot actions depend on different relative 3D relations at different stages of manipulation. For example, placing a cup into a bowl requires reasoning about gripper--cup relations during grasping and cup--bowl relations during placement, while other object pairs may be less relevant.
These relations change with robot and object motion, while geometric estimation errors, viewpoint changes, and occasional occlusion can make their current configurations difficult to determine reliably~\cite{peng2026spatialbench}.
SpatialVLA and 3DS-VLA improve spatial awareness through geometric representations and sequential spatial constraints~\cite{qu2025spatialvla,li2025threedsvla}. GraphCoT-VLA and TriRelVLA further explore structured 3D relations for manipulation~\cite{huang2026graphcotvla, zhou2026trirelvla}.
This raises a central question: How can VLA models explicitly model task-relevant 3D relations and continuously update their representations as manipulation unfolds to support action generation?

To address this challenge, we introduce \textbf{WeaveVLA}, a framework that explicitly models task-relevant relative 3D relations and their evolution during manipulation to enhance VLA action generation, as illustrated in Fig.~\ref{fig:teaser}.
Its object-centric representation combines visual features with estimated 3D geometry, using the robot's tool center point (TCP) to anchor gripper--object relations.
A task-conditioned graph connects instruction-relevant objects, the gripper, and nearby spatial references, while edge-aware graph attention encodes their relative 3D geometry.
To maintain evolving spatial context, WeaveVLA assigns a recurrent state to each selected entity pair and updates it with valid geometric evidence across observations.
The resulting relation features are conditioned on the vision-language context and summarized into compact relational tokens. An auxiliary geometry reconstruction objective further encourages the learned relational features to preserve spatial structure.
Together with the original vision-language representations, these tokens condition a continuous action expert for robot manipulation.

We instantiate WeaveVLA on two VLA backbones, StarVLA-GR00T and $\pi_{0.5}$, and evaluate its effectiveness on LIBERO, LIBERO-Plus, and CALVIN~\cite{mees2022calvin}.
On the four standard LIBERO suites, WeaveVLA achieves average success rates of 98.3\% and 98.2\%, respectively, improving both backbones.
On LIBERO-Plus, it improves StarVLA-GR00T from 74.9\% to 82.0\% and $\pi_{0.5}$ from 85.7\% to 87.8\%, demonstrating consistent gains across the two architectures.
On CALVIN D$\rightarrow$D, WeaveVLA increases the average completed sequence length from 3.645 to 3.910 for StarVLA-GR00T, surpassing the original $\pi_{0.5}$ baseline, and from 3.885 to 3.997 for $\pi_{0.5}$.
Across three real-world tasks, the StarVLA-GR00T-based WeaveVLA achieves average success rates of 70.0\% under standard conditions and 53.3\% under perturbations, outperforming both baselines.

Our main contributions are summarized as follows:
\begin{enumerate}
    \item We introduce WeaveVLA, a 3D relational grounding framework that explicitly models task-relevant spatial relations among objects, the gripper, and nearby references to enhance VLA action generation.

    \item We develop pair-specific recurrent relational updating to maintain evolving 3D relations across observations, integrating valid geometric evidence while preserving historical relational context.

    \item We demonstrate consistent improvements across two VLA backbones on LIBERO, LIBERO-Plus, and CALVIN D$\rightarrow$D, and validate WeaveVLA on three real-world manipulation tasks, outperforming both baselines under standard and OOD conditions.

\end{enumerate}

\begin{figure}[!t]
\centering
\includegraphics[width=\textwidth]{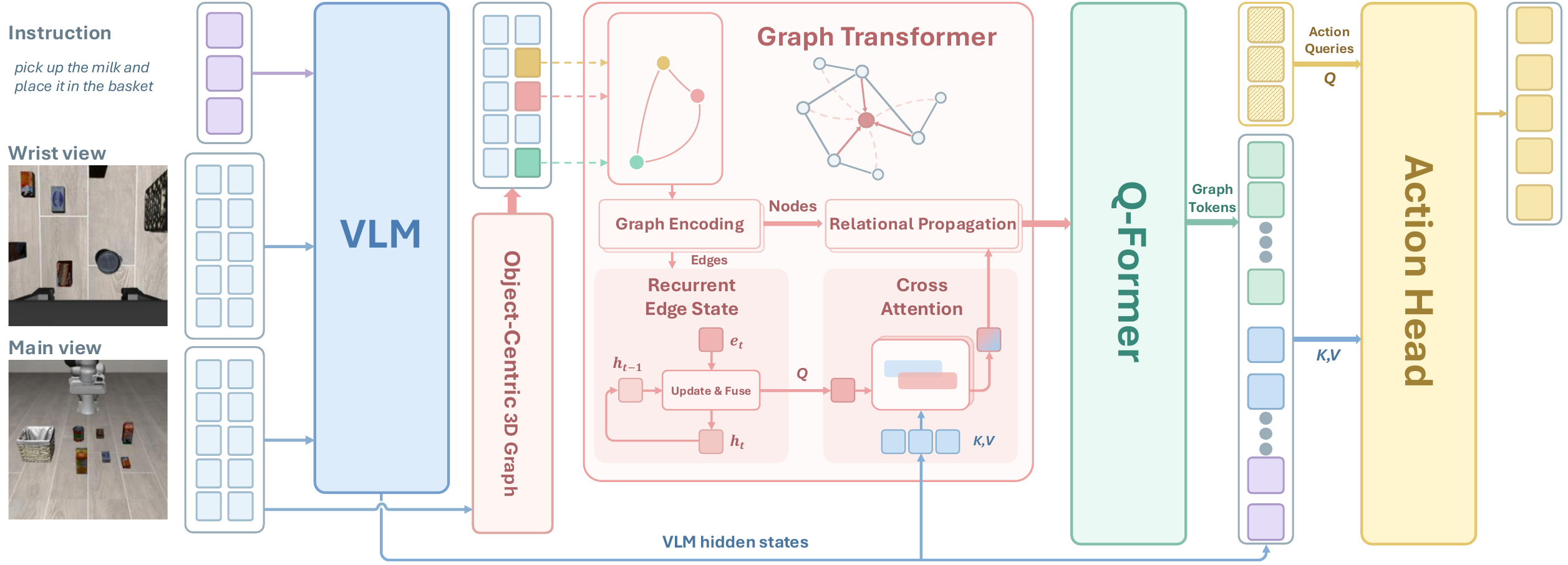}
\caption{Architecture of WeaveVLA. Object-level visual features and relative 3D geometry form a task-conditioned relational graph. Graph Transformer blocks encode and propagate relational features, while pair-specific recurrent edge states integrate valid geometric evidence across observations. Cross-attention incorporates vision-language context, and a Q-Former extracts compact relational tokens. The action interface shown is for StarVLA-GR00T.}
\label{fig:overview}
\end{figure}

\section{Related Work}
\label{sec:related}
\subsection{Vision-Language-Action Models}
Building on pretrained vision-language representations and robot demonstrations, vision-language-action (VLA) research has advanced toward more generalizable robot manipulation.
Early approaches such as RT-2 and OpenVLA demonstrated the transfer of visual-semantic knowledge to robotic control~\cite{brohan2023rt2,kim2024openvla}. OpenVLA-OFT improved action generation efficiency through continuous action representations, parallel decoding, and action chunking~\cite{kim2025fine}. 
Subsequent VLA architectures further advanced continuous action generation and manipulation across diverse tasks~\cite{black2024pi0,gr00t2025,pan2026worldtowrist}. More recent models have expanded generalization and adaptability, with $\pi_{0.5}$ demonstrating generalization to unseen environments~\cite{intelligence2025pi_} and $\pi_{0.7}$ advancing steerable execution and compositional generalization~\cite{intelligence2026pi07steerablegeneralistrobotic}. 
LingBot-VLA 2.0 and GR00T N1.7 further advanced cross-embodiment manipulation~\cite{wu2026lingbotvla2,nvidia2026gr00tn17}, while Gemini Robotics 2 extended generalist control to whole-body and dexterous manipulation. 
WeaveVLA complements these advances by augmenting VLA action generation with explicit, task-relevant 3D spatial relations.

\subsection{Spatial Grounding for Robotic Manipulation}

Spatial grounding connects visual observations with the geometric structure of the environment to support robotic manipulation. 
Recent studies have incorporated explicit 3D representations into VLA policies. SpatialVLA introduces 3D positional encoding to enhance spatial awareness~\cite{qu2025spatialvla}, while 3DS-VLA combines 2D-3D feature alignment with sequential spatial constraints for manipulation~\cite{li2025threedsvla}. 
Graph-based representations have also been explored for spatial relation modeling in navigation and autonomous driving, exemplified by BSG and GraphAD~\cite{liu2023bsg,zhang2024graphad}. Graph-of-Mark further enhances spatial reasoning for multimodal models via graph-based visual prompting~\cite{gom2026aaai}.

For robotic manipulation, scene graphs have been used for both high-level planning and action conditioning. 
Compose by Focus employs task-relevant graphs to condition diffusion-based manipulation skills~\cite{qi2026focus}, while GraSP-VLA uses scene graphs for symbolic planning and policy orchestration~\cite{neau2025graspvla}. 
SIR further learns task-relevant graph sparsification for action generation~\cite{mattes2026sir}. More directly, GraphCoT-VLA combines an updatable 3D pose-object graph with structured reasoning~\cite{huang2026graphcotvla}, while TriRelVLA models object-hand-task relations to condition action prediction~\cite{zhou2026trirelvla}. 
Other studies have explored spatiotemporal representations for manipulation: ConsisVLA-4D addresses spatiotemporal consistency~\cite{li2026consisvla}, semantic-geometric task representations jointly encode object relations and motion histories~\cite{herbert2026semantic}, and Qian et al. maintain dynamic scene graphs as structured memory under partial observability~\cite{qian2026scenegraphs}. 
While these studies demonstrate the value of spatial structure and temporal context, WeaveVLA focuses on explicitly encoding task-relevant relative 3D relations for action generation. A task-conditioned graph captures gripper--object and object--reference geometry, while observation-aware recurrent updates maintain pair-specific relational states across observations.

\section{Method}
\label{sec:method}

WeaveVLA grounds robot actions in task-relevant relative 3D relations, as illustrated in Fig.~\ref{fig:overview}.
Given RGB observations from the available cameras, proprioception $s_t$, and instruction $\ell$, the framework associates tracked entities with visual features and estimated 3D geometry.
A task-conditioned graph organizes relative spatial relations among relevant objects, the gripper, and nearby references, while pair-specific recurrent states maintain these relations across observations.
A learned-query readout then extracts compact relational tokens from the updated graph features, which complement the original vision-language context $H_t$ to condition a continuous action expert.

\subsection{Object-Centric Observation Representation}
\label{sec:perception}

To ground 3D spatial relations, WeaveVLA first establishes object-centric visual representations with consistent identities across observations.
At episode initialization, an adapted YOLOE detector~\cite{wang2025yoloe} supplies labeled boxes to frozen SAM~3~\cite{carion2025sam3}; Cutie~\cite{cheng2024cutie} then propagates instance masks and identities.
Robot regions are excluded from object instance masks before feature pooling. Object appearance features are pooled from visual tokens aligned with instance masks in the same camera view, while the original VLM retains its backbone-specific camera inputs.

Since visual tokens may only partially overlap an object, we aggregate instance features according to their foreground coverage.
For a visible instance $i$ with a nonempty current mask, let $a_{ip,t}$ denote its foreground area within visual token grid cell $p$, with feature $f_{p,t}$.
The instance appearance feature is computed as
\begin{align}
v_{i,t}
&= \sum_{p:a_{ip,t}>0}
\frac{a_{ip,t}}{\sum_q a_{iq,t}}f_{p,t}.
\label{eq:pooling}
\end{align}

The gripper is represented as a separate entity. Its finger masks are obtained from robot geometry and camera calibration, and appearance features are pooled from visible finger regions using the same foreground-coverage weighting. Its 3D position is anchored at the calibrated tool center point (TCP), obtained from robot kinematics rather than visual depth.
Detector adaptation, mask filtering, and tracking details appear in Appendix~\ref{app:perception}.

\subsection{Spatial Graph Construction and Encoding}
\label{sec:topology}

Spatial grounding aligns geometry of tracked objects and the gripper TCP within a shared image-aligned coordinate frame. Masked metric depth from DA3Metric-Large~\cite{lin2025da3} is backprojected using camera intrinsics, while the calibrated TCP is transformed into the same frame. Stable references help mitigate depth-scale drift under a fixed camera, with the correction applied only to estimated object geometry. Node features combine foreground appearance, averaged VLM input embeddings of the category label, and a point MLP with max pooling over centered point clouds. Appendix~\ref{app:extensions} details the geometry-retention rules and graph-attention operations.

The instruction and local geometry jointly determine graph connectivity. A parser and frozen sentence encoder match instruction phrases to object categories and bind the corresponding candidate instances once per episode. Task candidates connect to one another and to the gripper. 
Each candidate with current geometric support selects its
$K=2$ nearest currently supported objects, counting
existing connections and including unmentioned objects. 
Selected connections are bidirectional, and eligible object edges can persist during occlusion. Scene nodes with usable geometry can contribute to the readout even if they have no selected edges.

For each selected ordered entity pair, we explicitly encode
its relative 3D displacement and distance:
\begin{align}
g_{ij,t}
&= [(c_{j,t}-c_{i,t})^\top,
\|c_{j,t}-c_{i,t}\|_2]^\top,
\label{eq:geometry}
\end{align}
where $c_{i,t}$ denotes the 3D reference position of entity $i$.
We use the calibrated TCP for the gripper and the center of
accepted current or eligible retained geometry for an object.
The geometric descriptor is embedded into an initial edge feature $e_{ij,t}^{(0)}=\operatorname{MLP}_{\mathrm{geom}}(g_{ij,t})$.

We employ two stages of edge-aware graph attention
over the same task-conditioned graph.
The first stage encodes node features and relative
3D geometry, while the second stage propagates
relational features after recurrent updating and
VLM conditioning. For each node $i$, attention is computed over its selected neighbors $\mathcal A_i$, including itself. At each layer, edge features influence both the attention weights and the messages propagated between nodes. For attention head $r$, omitting time and layer indices:
\begin{align}
\alpha_{ij}^{r}
&=\operatorname{softmax}_{j\in\mathcal A_i}
\left(
\frac{(q_i^r)^\top k_j^r}{\sqrt{d_h}}
+b_r(e_{ij})
\right),
\label{eq:graphattention}
\\
\mu_i^r
&=\sum_{j\in\mathcal A_i}
\alpha_{ij}^r
\left(v_j^r+u_r(e_{ij})\right),
\label{eq:graphmessage}
\end{align}
where $q_i^r$, $k_j^r$, and $v_j^r$ are head-specific projections of node features, and $d_h$ is the per-head dimension,
while $b_r$ and $u_r$ project edge features into attention biases and message embeddings, respectively. Self-attention uses zero edge contributions without introducing physical self-edges. 
Messages from all heads are concatenated and passed through an output projection, followed by residual normalization, a residual feed-forward block, and a second normalization to update the node features.

Meanwhile, each directed edge is updated using its previous representation and the input features of its two endpoints:
\begin{align}
e_{ij}^{(\ell+1)}
&=\operatorname{LN}\left(
e_{ij}^{(\ell)}
+\phi_R\left(
h_i^{(\ell)}\cat h_j^{(\ell)}
\cat e_{ij}^{(\ell)}
\right)
\right),
\label{eq:graphedge}
\end{align}
where $\phi_R$ is an MLP and $\cat$ denotes concatenation. The resulting edge representation $\bar e_{ij,t}$ encodes relative geometry together with information from connected entities, providing the spatial input for recurrent relational updating.

\subsection{Relational Updating and Action Conditioning}
\label{sec:conditioning}

As manipulation progresses, a single entity may participate in multiple spatial interactions that evolve differently.
We therefore maintain a separate recurrent state for each selected ordered entity pair to capture its relational history. A shared gated recurrent unit (GRU) updates these states:
\begin{align}
m_{ij,t}
&=
\begin{cases}
\operatorname{GRU}(\bar e_{ij,t},m_{ij,t^-}),
& (i,j)\in\mathcal E_t^{\mathrm{obs}},\\
m_{ij,t^-}, & \text{otherwise},
\end{cases}
\label{eq:memory}
\end{align}
Here $t^-$ denotes the previous update. The set $\mathcal E_t^{\mathrm{obs}}$ contains edges with valid current geometric support at both endpoints. Such support can come from newly accepted object geometry, static geometry validated by the current observation, or the current TCP. Geometry retained solely during occlusion may still contribute to spatial encoding, but does not trigger recurrent updates. Edges without valid support at both endpoints retain their previous states, with updates resuming when valid support becomes available, subject to expiry and reset rules.

The maintained relational states are fused with spatially encoded edge features and conditioned on the vision-language context:
\begin{align}
\begin{aligned}
\widetilde e_{ij,t}
&=\phi_M(\bar e_{ij,t}\cat m_{ij,t}),\\
E_t^{\mathrm{cond}}
&=\Phi_C\left(\widetilde E_t+
\Attn(\widetilde E_t,H_t,H_t)\right).
\end{aligned}
\label{eq:relationcontext}
\end{align}
The operator $\cat$ denotes concatenation, $\phi_M$ is an MLP, and $\Phi_C$ is a normalized residual block. The second graph-attention stage propagates the VLM-conditioned edge features over the same task-conditioned graph, producing the final node and edge representations $U_t$ and $E_t$.

Learned queries $Q$ read the resulting node and edge representations to generate a fixed number of relational tokens:
\begin{align}
T_t^G
&=W_{\mathrm{out}}\operatorname{Readout}(Q,[U_t;E_t]).
\label{eq:tokens}
\end{align}
The queries are shared across tasks, and $W_{\mathrm{out}}$ maps the readout features to the token width required by each backbone's action expert.

For StarVLA-GR00T, we append the relational tokens to the vision-language hidden states to form the cross-attention context $C_t^{\mathrm{act}}=[H_t;T_t^G]$. For $\pi_{0.5}$, we prepend the projected relational tokens to the noisy action tokens in the action expert's input sequence. At each denoising step, the action tokens attend to both the relational tokens and the original vision-language prefix. Appendix~\ref{app:policyinference} details inference for both backbones.

\subsection{Training Objectives}
\label{sec:training}

The action expert learns continuous control through flow matching~\cite{lipman2023flow}. Given a demonstration action chunk $A_t^*$ and Gaussian noise $\epsilon\sim\mathcal N(0,I)$, we use the noise-to-action convention $A_t^\tau=(1-\tau)\epsilon+\tau A_t^*$. The interpolation time $\tau$ follows a backbone-specific training distribution, detailed in Appendix~\ref{app:training_protocol}. The action expert minimizes
\begin{align}
\mathcal L_{\mathrm{FM}}
&=\mathbb E\left[
\frac{\|v_\phi(A_t^\tau,\tau;H_t,T_t^G,s_t)
-(A_t^*-\epsilon)\|_F^2}{H d_a}\right],
\label{eq:flowloss}
\end{align}
where $H$ and $d_a$ denote the action chunk length and dimension. To preserve relative 3D geometry through graph encoding and VLM conditioning, an auxiliary head reconstructs descriptor~\eqref{eq:geometry} from the final edge features $E_t$. The smooth-$L_1$ loss $\mathcal L_{\mathrm{geo}}$ uses detached geometric targets on valid observed edges. The joint training objective is
\begin{align}
\mathcal L
&=\mathcal L_{\mathrm{FM}}+\lambda_g\mathcal L_{\mathrm{geo}}.
\label{eq:baseloss}
\end{align}

Here, $\lambda_g$ weights the auxiliary geometry loss. During policy training, the external perception modules remain frozen, and training and inference share the same TCP-anchored, coverage-weighted node construction.
Recurrent states persist across observations within an episode but are detached between successive updates, preserving relational history while restricting gradient propagation to the current recurrent update.

\section{Experiments}
\label{sec:experiments}
We evaluate how explicit 3D relational grounding enhances StarVLA-GR00T and $\pi_{0.5}$ on LIBERO~\cite{liu2023libero}, LIBERO-Plus~\cite{fei2026liberoplus}, and CALVIN~\cite{mees2022calvin}, comparing each WeaveVLA instantiation with its corresponding backbone. We further evaluate the StarVLA-GR00T-based variant on three real-world manipulation tasks and analyze key design choices and inference latency.

\subsection{Experimental Setup}
\label{sec:experimentalsetup}
Dashes denote unavailable results throughout the tables. In benchmark comparisons, red/bold and blue/underlined values indicate the best and second-best scores, including ties.

\noindent\textbf{Implementation details.}
We instantiate WeaveVLA on two VLA backbones, StarVLA-GR00T~\cite{starvla2026} and $\pi_{0.5}$~\cite{intelligence2025pi_}. The StarVLA-GR00T instantiation uses Qwen3-VL-4B-Instruct~\cite{bai2025qwen3vl} and the QwenGR00T action framework. The relational module has a hidden width of 256 and eight attention heads, with two layers each for spatial encoding, VLM-conditioned graph propagation, and learned-query readout. We use 16 relational tokens by default. Backbone-specific training configurations and hyperparameters for each benchmark are provided in Appendix~\ref{app:training}.

\smallskip
\noindent\textbf{LIBERO.}
LIBERO comprises four suites: Spatial, Object, Goal, and Long, with Long corresponding to LIBERO-10. We evaluate 500 rollouts per suite and report the success rate of each suite and their average.

\smallskip
\noindent\textbf{LIBERO-Plus.}
LIBERO-Plus contains 10,030 perturbed tasks spanning seven perturbation categories. For each model, we reuse its LIBERO-trained checkpoint with unchanged inference settings, without additional training or fine-tuning on LIBERO-Plus. Each perturbed task is evaluated once, and overall success is pooled across all 10,030 trials. Both benchmarks use control-step limits of 220, 280, 300, and 520 for the Spatial, Object, Goal, and Long suites, respectively, with ten settling steps.

\smallskip
\noindent\textbf{CALVIN.}
We evaluate WeaveVLA on CALVIN~\cite{mees2022calvin} under the D$\rightarrow$D setting, where models are trained and evaluated in environment D using separate training and evaluation sequences. The evaluation consists of 1,000 sequences of five consecutive language-conditioned subtasks, with at most 360 control steps per subtask.
We report the success rates for completing at least one through five consecutive subtasks and the average completed sequence length (Avg.~Len.).
Each instruction is treated as a separate graph-state episode, with task bindings and recurrent edge states reset at each subtask transition.

\begin{figure}[!t]
\centering
\includegraphics[width=\textwidth]{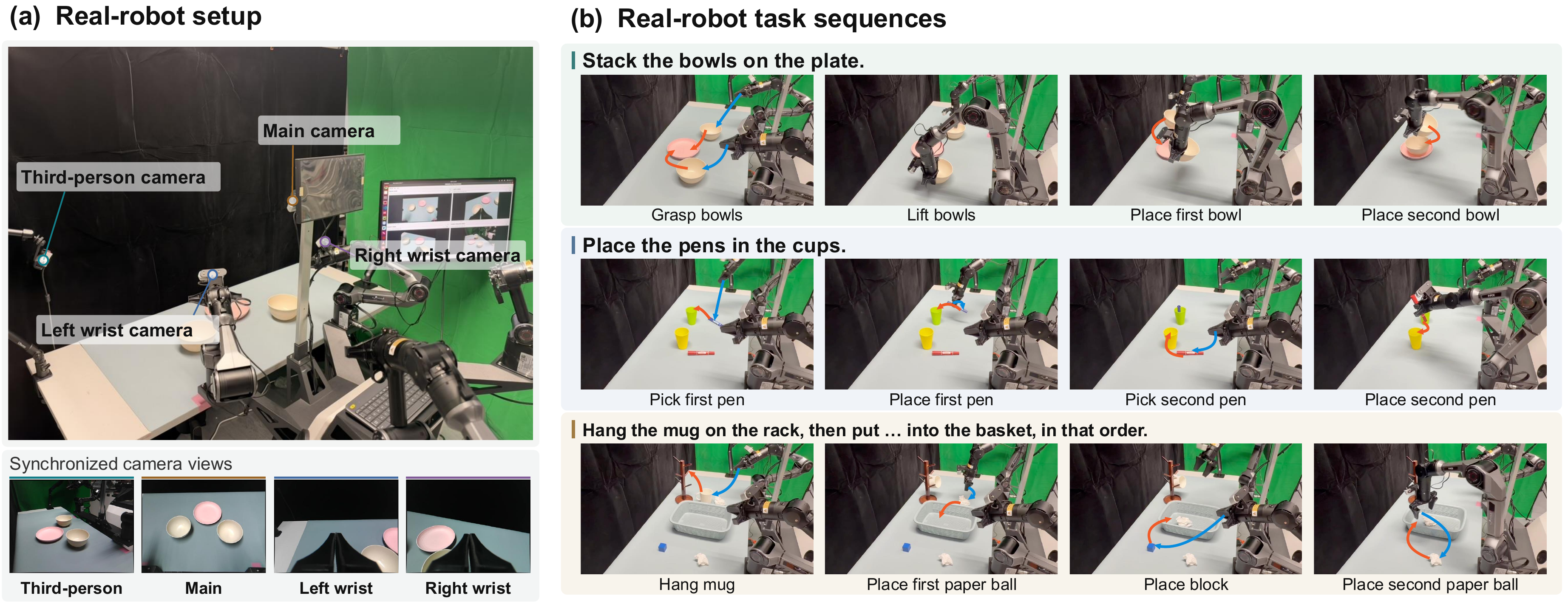}
\caption{Real-robot setup and example task trajectories. (a) The dual-arm platform uses a third-person camera for graph construction, a high-mounted main camera, and two wrist cameras. (b) Recorded demonstrations of stacking bowls, placing pens in cups, and cleaning the table. Each row shows four chronological frames from a single demonstration.}
\label{fig:real_robot}
\end{figure}

\smallskip
\noindent\textbf{Real-world setup.}
We evaluate three real-world manipulation tasks as shown in Fig.~\ref{fig:real_robot}.
\textit{1) Bowl Stacking.} The robot places the first bowl on a plate, then uses its position to align and stack the second bowl.
\textit{2) Place Pens in Cups.} The robot sequentially picks up two pens and places each in its designated cup.
\textit{3) Clean Table.} The robot hangs the mug on the rack, then places the first paper ball, the block, and the second paper ball in the basket.
For each task and method, we conduct 30 trials under standard conditions and 30 out-of-distribution (OOD) trials, evenly split across table clutter, lighting perturbations, and background variations. We report binary task success and a stage-wise progress score based on completed stages in the predefined task sequence. Appendix~\ref{app:realworld-progress} provides detailed task configurations, OOD settings, and progress scoring criteria.

\providecommand{\gainup}[1]{\rlap{\raisebox{-0.5ex}[0pt][0pt]{\scriptsize\textcolor{graphgreen}{(+#1)}}}}

\newcommand{\liberoscoregain}[2]{\shortstack[c]{#1\\[0.5pt]{\tiny\textcolor{graphgreen}{(+#2)}}}}

\begin{table}[t]
\centering
\caption{Comparison on LIBERO. Success rates (\%) are reported across four suites, with Long denoting LIBERO-10. Avg. is the four-suite mean. Green values indicate percentage-point gains over the corresponding backbone.}
\label{tab:liberoresults}
{\papertablesetup
\setlength{\tabcolsep}{2pt}
\renewcommand{\arraystretch}{1.10}
\begin{tabular*}{0.98\linewidth}{@{\extracolsep{\fill}}lccccc@{}}
\toprule
\tablehead{Method}
& \tablehead{Spat.}
& \tablehead{Obj.}
& \tablehead{Goal}
& \tablehead{Long}
& \tablehead{Avg.}\\
\midrule

OpenVLA~\cite{kim2024openvla}
& 84.7 & 88.4 & 79.2 & 53.7 & 76.5\\

OpenVLA-OFT~\cite{kim2025fine}
& 97.6 & 98.4 & 97.9 & 94.5 & 97.1\\

$\pi_0$~\cite{black2024pi0}
& 96.8 & 98.8 & 95.8 & 85.2 & 94.2\\

DreamVLA~\cite{zhang2026dreamvla}
& 97.5 & 94.0 & 89.5 & 89.5 & 92.6\\

GR00T-N1~\cite{gr00t2025}
& 94.4 & 97.6 & 93.0 & 90.6 & 93.9\\

MemoryVLA~\cite{shi2025memoryvla}
& 98.4 & 98.4 & 96.4 & 93.4 & 96.7\\

VLA-JEPA~\cite{sun2026vla}
& 96.2 & \resultbest{99.6} & 97.2 & 95.8 & 97.2\\

DeepThinkVLA~\cite{yin2025deepthinkvla}
& 96.6 & 99.0 & 96.4 & \resultsecond{96.2} & 97.0\\

StarVLA-GR00T~\cite{starvla2026}
& 97.8 & 98.8 & 97.4 & 92.0 & 96.5\\

$\pi_{0.5}$~\cite{intelligence2025pi_}
& \resultsecond{98.8}
& 98.2
& 98.0
& 92.4
& 96.9\\

\midrule

\raisebox{1ex}[0pt][0pt]{WeaveVLA (StarVLA)}
& \liberoscoregain{\resultsecond{98.8}}{1.0}
& \liberoscoregain{\resultbest{99.6}}{0.8}
& \liberoscoregain{\resultbest{98.8}}{1.4}
& \liberoscoregain{96.0}{4.0}
& \liberoscoregain{\resultbest{98.3}}{1.8}\\

\addlinespace[4pt]

\raisebox{1ex}[0pt][0pt]{WeaveVLA ($\pi_{0.5}$)}
& \liberoscoregain{\resultbest{99.0}}{0.2}
& \liberoscoregain{\resultsecond{99.2}}{1.0}
& \liberoscoregain{\resultsecond{98.2}}{0.2}
& \liberoscoregain{\resultbest{96.4}}{4.0}
& \liberoscoregain{\resultsecond{98.2}}{1.3}\\

\addlinespace[2pt]
\bottomrule
\end{tabular*}}
\end{table}

\subsection{Simulation Results}
\label{sec:simulation}

\providecommand{\gainup}[1]{\rlap{\raisebox{-0.5ex}[0pt][0pt]{\scriptsize\textcolor{graphgreen}{(+#1)}}}}
\newcommand{\gaindown}[1]{\rlap{\raisebox{-0.5ex}[0pt][0pt]{\scriptsize(-#1)}}}

\begin{table}[t]
\centering
\caption{Zero-shot evaluation on LIBERO-Plus using LIBERO-trained models. Success rates (\%) are reported across seven perturbation categories, with overall success pooled over all 10,030 tasks. StarVLA-GR00T results are from our reproduction. Parenthesized values indicate percentage-point changes relative to the corresponding backbone, with positive gains highlighted in green.}
\label{tab:pluscomparison}
{\papertablesetup
\setlength{\tabcolsep}{3pt}
\begin{tabular*}{0.98\linewidth}{@{\extracolsep{\fill}}l|cccccccc@{\hspace{1em}}}
\toprule
\tablehead{Method}
& \tablehead{Camera}
& \tablehead{Robot}
& \tablehead{Language}
& \tablehead{Light}
& \tablehead{Background}
& \tablehead{Noise}
& \tablehead{Layout}
& \tablehead{Overall}\\
\midrule

Fast-WAM~\cite{yuan2026fastwam}
& 16.4 & 44.5 & 68.9 & 78.2 & 53.7 & 37.7 & 60.7 & 51.5\\

$\pi_0$~\cite{black2024pi0}
& 13.8 & 6.0 & 58.8 & 85.0 & 81.4 & 79.0 & 68.9 & 53.6\\

VLA-JEPA~\cite{sun2026vla}
& 63.3 & 67.1 & 85.4 & 95.6 & 93.6 & 66.3 & 85.1 & 79.5\\

OpenVLA-OFT~\cite{kim2025fine}
& 56.4 & 31.9 & 79.5 & 88.7 & 93.3 & 75.8 & 74.2 & 69.6\\

Being-H0.7~\cite{luo2026being}
& \resultbest{82.0}
& 59.0
& 82.8
& 97.8
& 90.0
& \resultsecond{93.5}
& \resultbest{88.5}
& 82.1\\

Cosmos-Policy~\cite{kim2026cosmospolicy}
& 75.8 & 63.3 & 81.7 & 96.5 & 88.9 & 92.7 & 82.2 & 82.2\\

ImageWAM~\cite{zhang2026imagewam}
& \resultsecond{80.8}
& 50.3
& \resultsecond{91.4}
& \resultbest{98.1}
& 85.5
& \resultbest{93.8}
& 80.5
& 83.1\\

ABot-M0.5~\cite{chen2026abotm05}
& 70.5
& \resultbest{87.4}
& 88.6
& 94.0
& 89.7
& 75.5
& 85.2
& 83.4\\

StarVLA-GR00T~\cite{starvla2026}
& 49.2
& 62.3
& 84.3
& 93.2
& 92.7
& 74.2
& 79.5
& 74.9\\

$\pi_{0.5}$~\cite{intelligence2025pi_}
& 75.4
& 77.5
& 85.6
& 96.9
& 94.6
& 89.7
& 85.7
& \resultsecond{85.7}\\

\midrule

WeaveVLA (StarVLA)
& 64.6\gainup{15.4}
& 69.2\gainup{6.9}
& \resultbest{93.2}\gainup{8.9}
& 94.8\gainup{1.6}
& \resultsecond{95.3}\gainup{2.6}
& 83.6\gainup{9.4}
& 81.2\gainup{1.7}
& 82.0\gainup{7.1}\\

WeaveVLA ($\pi_{0.5}$)
& 79.5\gainup{4.1}
& \resultsecond{85.7}\gainup{8.2}
& 83.1\gaindown{2.5}
& \resultsecond{97.9}\gainup{1.0}
& \resultbest{97.4}\gainup{2.8}
& 89.3\gaindown{0.4}
& \resultsecond{87.7}\gainup{2.0}
& \resultbest{87.8}\gainup{2.1}\\

\bottomrule
\end{tabular*}}
\end{table}

\subsubsection{LIBERO}
\label{sec:liberoresults}

Table~\ref{tab:liberoresults} compares both WeaveVLA instantiations with existing VLA models on LIBERO. With StarVLA-GR00T as the backbone, WeaveVLA achieves the highest average success rate of 98.3\%, improving upon its baseline by 1.8 percentage points. The $\pi_{0.5}$-based variant achieves a comparable average success rate of 98.2\%, outperforming its baseline by 1.3 percentage points. Both instantiations improve upon their corresponding backbones across all four suites, with a 4.0-percentage-point gain on LIBERO-Long in each case, highlighting the benefits of explicit relational grounding for long-horizon manipulation.

\subsubsection{LIBERO-Plus}
\label{sec:liberoplus}

Table~\ref{tab:pluscomparison} compares WeaveVLA instantiations with their corresponding backbones on LIBERO-Plus. WeaveVLA improves pooled success from 74.9\% to 82.0\% with StarVLA-GR00T and from 85.7\% to 87.8\% with $\pi_{0.5}$, corresponding to gains of 7.1 and 2.1 percentage points, respectively.

Notably, the StarVLA-based variant improves performance across all seven perturbation categories, with substantial gains of 15.4 points under Camera, 9.4 under Noise, 8.9 under Language, and 6.9 under Robot perturbations. The $\pi_{0.5}$-based variant achieves gains of 8.2 and 4.1 points under Robot and Camera perturbations, respectively, while reaching 97.9\% and 97.4\% success under Light and Background perturbations. These results demonstrate the effectiveness of explicit 3D relational grounding across diverse perturbations and VLA backbones.

\newcommand{\calvingain}[1]{{\scriptsize\textcolor{graphgreen}{(+#1)}}}

\begin{table}[t]
\centering
\caption{CALVIN D$\rightarrow$D evaluation on 1,000 five-instruction sequences. Columns 1--5 report success rates (\%) for completing at least the corresponding number of consecutive subtasks. Avg.~Len. denotes the average completed sequence length. Parenthesized values indicate changes relative to the corresponding backbone, with positive gains highlighted in green.}
\label{tab:calvincomparison}
{\papertablesetup
\setlength{\tabcolsep}{2pt}
\begin{tabular*}{0.98\linewidth}{@{\extracolsep{\fill}}l|cccccc@{}}
\toprule
\tablehead{Method}
& \tablehead{1}
& \tablehead{2}
& \tablehead{3}
& \tablehead{4}
& \tablehead{5}
& \tablehead{Avg.~Len.}\\
\midrule

RoboFlamingo~\cite{li2024vision}
& 86.0 & 71.4 & 58.5 & 46.0 & 34.9 & 2.968\\

MDT-V~\cite{reuss2024multimodal}
& \resultsecond{93.9}
& 83.8 & 73.5 & 63.9 & 54.9 & 3.700\\

RoboUniView~\cite{li2024RoboUniView}
& \resultbest{96.2}
& \resultbest{88.8}
& 77.6
& 66.6 & 56.3 & 3.855\\

StarVLA-GR00T~\cite{starvla2026}
& 89.8 & 79.9 & 71.8
& 64.9 & 58.1 & 3.645\\

$\pi_{0.5}$~\cite{intelligence2025pi_}
& 92.5 & 84.0 & 76.6
& 71.0 & 64.4 & 3.885\\

\midrule

\multirow{2}{*}{WeaveVLA (StarVLA)}
& 92.6
& \resultsecond{84.4}
& \resultsecond{77.9}
& \resultsecond{71.4}
& \resultsecond{64.7}
& \resultsecond{3.910}\\[-0.2ex]
&
\calvingain{2.8}
& \calvingain{4.5}
& \calvingain{6.1}
& \calvingain{6.5}
& \calvingain{6.6}
& \calvingain{0.265}\\

\addlinespace[0.6ex]

\multirow{2}{*}{WeaveVLA ($\pi_{0.5}$)}
& 92.7
& \resultsecond{84.4}
& \resultbest{78.9}
& \resultbest{74.5}
& \resultbest{69.2}
& \resultbest{3.997}\\[-0.2ex]
&
\calvingain{0.2}
& \calvingain{0.4}
& \calvingain{2.3}
& \calvingain{3.5}
& \calvingain{4.8}
& \calvingain{0.112}\\

\bottomrule
\end{tabular*}}
\end{table}

\subsubsection{CALVIN}
\label{sec:calvinresults}

Table~\ref{tab:calvincomparison} compares WeaveVLA with existing methods on CALVIN D$\rightarrow$D. With StarVLA-GR00T as the backbone, WeaveVLA improves the average completed sequence length from 3.645 to 3.910, surpassing the original $\pi_{0.5}$ baseline. With $\pi_{0.5}$ as the backbone, WeaveVLA improves the average length from 3.885 to 3.997, achieving the highest result among the compared methods. The two WeaveVLA instantiations achieve five-subtask success rates of 64.7\% and 69.2\%, improving their corresponding backbones by 6.6 and 4.8 percentage points, respectively. These results demonstrate consistent improvements in sequential manipulation across both VLA backbones.

\subsection{Real-World Results}
\label{sec:realworld}

\begin{figure}[!b]
\centering
\includegraphics[width=\linewidth]{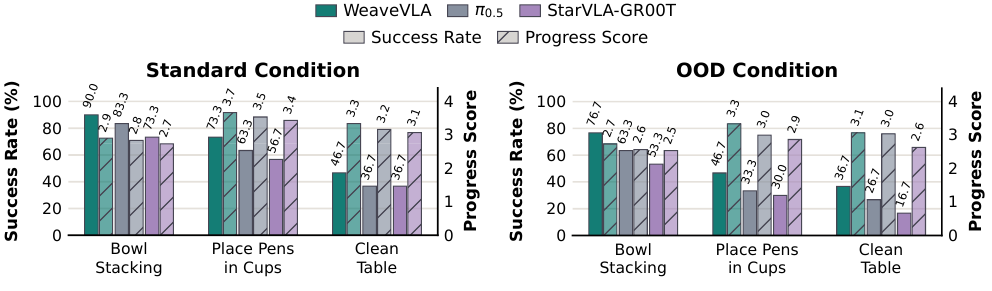}
\caption{Real-world results under standard and OOD conditions. Solid bars show success rates (left axis), while hatched bars show mean progress scores (right axis), with maximum progress scores of 3 for bowl stacking and 4 for the other tasks.}
\label{fig:realworld-results}
\end{figure}

WeaveVLA builds upon the StarVLA-GR00T backbone~\cite{starvla2026}. Figure~\ref{fig:realworld-results} compares its performance with StarVLA-GR00T and $\pi_{0.5}$~\cite{intelligence2025pi_} across three real-world tasks.

\smallskip
\noindent\textbf{Standard conditions.}
WeaveVLA achieves an average success rate of 70.0\%, improving on $\pi_{0.5}$ and StarVLA-GR00T by 8.9 and 14.4 percentage points, respectively. It succeeds in 90.0\% of bowl-stacking trials and 73.3\% of pen-placement trials, while table cleaning remains the most challenging task at 46.7\%.

\smallskip
\noindent\textbf{OOD conditions.}
WeaveVLA achieves an average success rate of 53.3\%, compared with 41.1\% for $\pi_{0.5}$ and 33.3\% for StarVLA-GR00T. Notably, although the original StarVLA-GR00T underperforms $\pi_{0.5}$, its WeaveVLA-enhanced variant surpasses $\pi_{0.5}$ under both conditions. It improves success across all three tasks, with the largest gain over StarVLA-GR00T on bowl stacking. WeaveVLA also achieves the highest mean progress score on each task under both conditions. On OOD pen placement, it completes 3.3 of four stages on average, compared with 3.0 for $\pi_{0.5}$ and 2.9 for StarVLA-GR00T.

Across the three tasks, WeaveVLA achieves 46.7\% success under table clutter, 50.0\% under random lighting, and 63.3\% under background variation. Its gains over $\pi_{0.5}$ are 3.3 percentage points under clutter and 16.7 points under each of the other perturbations. Table~\ref{tab:realworld-ood} in the appendix reports the breakdown by task and method.

\subsection{Ablation Studies and Further Analysis}
\label{sec:ablations}

\subsubsection{Ablations on LIBERO and LIBERO-Plus}
\label{sec:policytests}

Table~\ref{tab:policy} compares full WeaveVLA with three ablated
variants under matched training data, optimization budgets, and
evaluation conditions. Removing geometry supervision sets
$\lambda_g=0$ while retaining geometric inputs. Removing edge memory
disables the learned pair-specific states while preserving geometry
caching and action conditioning. The full-connectivity variant
connects all distinct valid scene nodes in both directions.
Appendix~\ref{app:policyablations} details the experimental controls.

Removing $\mathcal L_{\mathrm{geo}}$ reduces average success from 98.3\% to 97.7\% on LIBERO and from 82.0\% to 80.4\% on LIBERO-Plus, while removing edge memory yields 97.8\% and 80.7\%, respectively. Both components have a larger impact on LIBERO-Plus, suggesting that geometry supervision and recurrent relational memory contribute to robustness under perturbations. Replacing task-conditioned connections with full connectivity reduces performance to 97.5\% on LIBERO and 79.8\% on LIBERO-Plus, supporting the effectiveness of task-conditioned graph construction.

\begin{table}[t]
\centering
\caption{Ablation studies of WeaveVLA (StarVLA) on LIBERO and LIBERO-Plus. S, O, G, and L denote the four LIBERO suites, Avg. their mean, and Plus the pooled LIBERO-Plus success rate. All values are percentages.}
\label{tab:policy}
{\papertablesetup
\setlength{\tabcolsep}{2pt}
\begin{tabular*}{0.98\linewidth}{@{\extracolsep{\fill}}l|cccccc@{}}
\toprule
\tablehead{Variant}
& \tablehead{S $\uparrow$}
& \tablehead{O $\uparrow$}
& \tablehead{G $\uparrow$}
& \tablehead{L $\uparrow$}
& \tablehead{Avg. $\uparrow$}
& \tablehead{Plus $\uparrow$}\\
\midrule
StarVLA
& 97.8 & 98.8 & 97.4 & 92.0 & 96.5 & 74.9\\
\midrule
\textbf{WeaveVLA (Full)}
& 98.8 & 99.6 & 98.8 & 96.0 & \textbf{98.3} & \textbf{82.0}\\
w/o $\mathcal L_{\mathrm{geo}}$
& 98.6 & 99.8 & 98.4 & 94.0 & 97.7 & 80.4\\
w/o Edge Memory
& 98.4 & 98.6 & 98.4 & 95.6 & 97.8 & 80.7\\
w/ Full Connectivity
& 98.6 & 99.0 & 96.4 & 95.8 & 97.5 & 79.8\\
\bottomrule
\end{tabular*}}
\end{table}

\subsubsection{Error Analysis of Relative Geometry}
\label{sec:relative-geometry-error}

We analyze geometric estimation errors for absolute object positions and pairwise relative displacements using selected LIBERO-Plus trajectories. Both are computed from identical depth predictions and instance masks.

\begin{figure}[t]
    \centering
    \includegraphics[width=\linewidth]
    {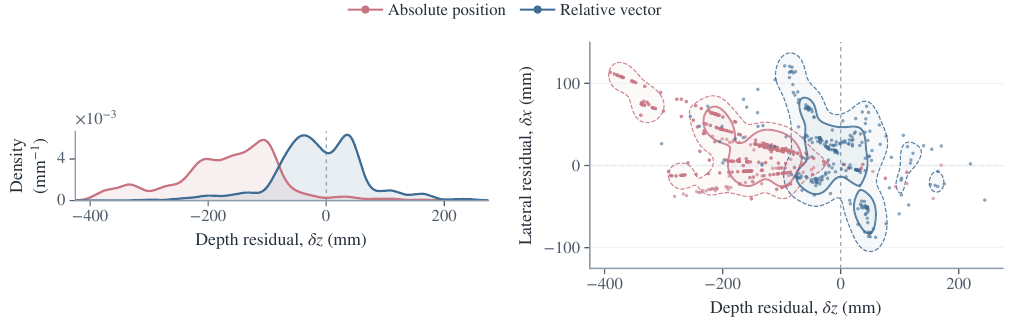}
    \caption{Camera-frame geometric errors for absolute
    object positions (rose) and relative displacement
    vectors (blue), using identical depth predictions
    and masks. Left: normalized distributions of signed
    depth residuals. Right: depth--lateral residuals
    with 50\% and 80\% density contours.}
    \label{fig:camera-error-cloud}
\end{figure}

As shown in Fig.~\ref{fig:camera-error-cloud}, relative displacements exhibit substantially lower geometric errors, reducing the RMSE by 53.6\%. This reduction is consistent with the partial cancellation of shared localization biases between objects, making relative geometry less sensitive to common position errors without requiring more accurate depth estimates.

\subsubsection{Inference Latency}
\label{sec:latency}

Table~\ref{tab:latency} compares inference latency with and without the WeaveVLA relational grounding pipeline for StarVLA-GR00T and $\pi_{0.5}$. Graph-off and graph-on configurations use the same model weights. Policy prediction measures the complete \texttt{predict\_action} call, while client response latency additionally includes online perception, geometric processing, communication, and action-response parsing. WeaveVLA increases policy prediction latency by 12.0 ms for StarVLA-GR00T and 3.6 ms for $\pi_{0.5}$, while client response latency increases by 63.1 and 48.9 ms, respectively. Most of the additional latency arises outside the policy call, primarily from online perception, geometric processing, and data handling.
The additional latency remains manageable, supporting real-time robot control.

\begin{table}[t]
\centering
\begin{minipage}{0.62\linewidth}
\centering
\caption{Inference latency on a single NVIDIA RTX 4090 GPU. Values are mean steady-state latencies (ms) with batch size one.}
\label{tab:latency}
{\papertablesetup
\begin{tabular*}{\linewidth}{@{\extracolsep{\fill}}l|cc@{}}
\toprule
\tablehead{Method} &
\shortstack{\tablehead{Policy}\\\tablehead{prediction (ms)}} &
\shortstack{\tablehead{Client}\\\tablehead{response (ms)}}\\
\midrule
StarVLA-GR00T (graph off) & 46.9 & 49.8\\
WeaveVLA (StarVLA) & 58.9 & 112.9\\
\midrule
$\pi_{0.5}$ (graph off) & 105.6 & 108.2\\
WeaveVLA ($\pi_{0.5}$) & 109.2 & 157.1\\
\bottomrule
\end{tabular*}}
\end{minipage}
\end{table}

\FloatBarrier

\section{Conclusion}
We have introduced WeaveVLA, a framework that enhances VLA models by explicitly encoding task-relevant relative 3D relations and modeling their evolution throughout manipulation. 
By combining task-conditioned sparse graph construction with pair-specific recurrent relational updating, WeaveVLA integrates current spatial geometry and historical relational context for continuous action generation.
Extensive evaluations on simulation benchmarks demonstrate consistent improvements across different VLA backbones, highlighting the effectiveness of explicit relational grounding for robotic manipulation. 
On three real-world tasks, WeaveVLA improves success over both baselines under standard and perturbed conditions.

\newpage
\bibliography{main}
\clearpage
\appendix
\addtocontents{toc}{\protect\setcounter{tocdepth}{1}}
\setcounter{tocdepth}{1}
\begingroup
\renewcommand{\contentsname}{Appendix}
\pdfbookmark[0]{Appendix}{appendix-contents}
\tableofcontents
\endgroup

\medskip
\section{Training Configuration}
\label{app:training}

\begin{minipage}{\textwidth}

Table~\ref{tab:liberotraining} summarizes the training configurations of WeaveVLA with StarVLA and $\pi_{0.5}$ backbones. On LIBERO, both variants use the same demonstration set. For CALVIN D$\rightarrow$D, both variants train on the training split of environment D. For all three real-world tasks, WeaveVLA uses the StarVLA-GR00T backbone and predicts absolute joint configurations from real-world demonstrations.

\par\medskip

For all three real-world tasks, the VLM receives four synchronized RGB views from the third-person camera, the high-mounted main camera, and both wrist cameras. We construct the graph from the third-person view and pool node appearance features from VLM tokens aligned with the instance masks in that view.

\par\medskip

\captionsetup{hypcap=false}
\captionof{table}{Training configurations of WeaveVLA on LIBERO, CALVIN, and real-world manipulation tasks.}
\label{tab:liberotraining}
\centering
{\papertablesetup
\begin{tabular*}{\linewidth}{@{\extracolsep{\fill}}lccccc@{}}
\toprule
& \multicolumn{2}{c}{\tablehead{LIBERO}}
& \multicolumn{2}{c}{\tablehead{CALVIN D$\rightarrow$D}}
& \tablehead{Real-world Tasks}\\
\cmidrule(lr){2-3}
\cmidrule(lr){4-5}
\cmidrule(l){6-6}
\tablehead{Setting}
& \shortstack{WeaveVLA\\(StarVLA)}
& \shortstack{WeaveVLA\\($\pi_{0.5}$)}
& \shortstack{WeaveVLA\\(StarVLA)}
& \shortstack{WeaveVLA\\($\pi_{0.5}$)}
& \shortstack{WeaveVLA\\(StarVLA)}\\
\midrule
Training data
& \multicolumn{2}{c}{1,693 demonstrations}
& \multicolumn{2}{c}{Environment D}
& 72 train / 9 validation demos\\
GPUs & 4 & 6 & 4 & 4 & 2\\
Global batch size & 96 & 256 & 32 & 32 & 48\\
Action dimensions & 7 & 7 & 7 & 7 & 14 (absolute joints)\\
Prediction horizon (steps) & 8 & 10 & 8 & 8 & 16\\
\midrule
Optimizer & AdamW & AdamW & AdamW & AdamW & AdamW\\
Backbone learning rate
& $10^{-5}$ & $5\times10^{-5}$
& $10^{-5}$ & $2.5\times10^{-5}$ & $10^{-5}$\\
Graph and action learning rate
& $10^{-4}$ & $5\times10^{-5}$
& $10^{-4}$ & $2.5\times10^{-5}$ & $10^{-4}$\\
Warm-up (updates)
& 5,000 & 10,000 & 5,000 & 5,000 & 1,000\\
Weight decay
& $10^{-8}$ & $10^{-10}$
& $10^{-8}$ & $10^{-8}$ & $10^{-8}$\\
Adam $(\beta_1,\beta_2)$
& $(0.9,0.95)$ & $(0.9,0.95)$
& $(0.9,0.95)$ & $(0.9,0.95)$
& $(0.9,0.95)$\\
Adam $\epsilon$
& $10^{-8}$ & $10^{-8}$
& $10^{-8}$ & $10^{-8}$ & $10^{-8}$\\
Forward precision
& BF16 & BF16 & BF16 & BF16 & BF16\\
Recurrent gradient window
& 1 update & 1 update & 1 update & 1 update & 1 update\\
Geometry loss weight $\lambda_g$
& 0.1 & 0.1 & 0.1 & 0.1 & 0.1\\
\bottomrule
\end{tabular*}}

\end{minipage}
\par\medskip

\section{Implementation Details}
\label{app:extensions}

WeaveVLA maintains graph states using tracked instance
identities and ordered entity pairs. Newly observed node
features replace cached values, while eligible historical
features are retained when observations are missing.
A shared GRU maintains separate recurrent states for each
ordered pair, initialized to zero for unseen pairs.
These states persist across observations but are detached
between updates. They are reset when the geometric reference changes, and expire
64 control steps after the last valid update.

Geometric validity is determined by observation consistency
and the age of retained information. Static-geometry consistency is assessed using mask IoU,
mask-centroid displacement, relative mask-area change,
and mean absolute RGB difference within an eroded common
mask region, with respective thresholds of 0.85,
1.5 pixels at a 224-pixel reference scale, 15\%,
and 8 intensity units.

Depth-scale correction requires two confirmations and at
least two reliable reference objects. The correction factor
is clipped to $[0.8,1.25]$ and can be retained for up to
16 control steps. Occluded object geometry may remain
available for 64 control steps under fixed calibration,
subject to clearance from the gripper.
The TCP remains independent of depth-scale correction.
Retained geometry can contribute to spatial encoding,
but only valid geometric support permits recurrent updates,
as defined in Eq.~\eqref{eq:memory}.

Instance appearance features are obtained through
foreground-coverage pooling, as described in
Eq.~\eqref{eq:pooling}. Training and inference use
identical pooling operations on backbone-specific
visual token grids. For StarVLA, Qwen3-VL resizes $224\times224$ images to
$256\times256$, yielding an $8\times8$ post-merger token grid,
while $\pi_{0.5}$ uses a $16\times16$ SigLIP patch grid.
Masks are aligned with their visual tokens, and
foreground coverage determines the pooling weights.
When instance masks or visible finger regions are unavailable,
the corresponding current visual features are set to zero,
while eligible historical node features are retained.

The initial node representation combines category
semantics, local point-cloud geometry, and pooled
appearance features:
\begin{align}
h_{i,t}^{(0)}
&=\phi_V\left(
W_\ell s_i^\ell
\cat p_{i,t}
\cat W_v v_{i,t}
\right),
\label{eq:node}
\end{align}
where $s_i^\ell$ denotes the category embedding,
$p_{i,t}$ the point-cloud feature, and $v_{i,t}$
the pooled visual feature. Here $\cat$ denotes
concatenation and $\phi_V$ is an MLP.

Graph connectivity follows the task-conditioned
construction described in Sec.~\ref{sec:topology}.
Each currently supported task object selects up to
$K=2$ nearest objects as local references.
Existing task connections count toward this limit,
and duplicate edges do not introduce additional
neighbors. Selected relations are bidirectional,
while eligible historical edges may persist during
temporary occlusion.

Both graph-attention stages use edge-dependent attention
biases and message embeddings, as defined in
Eqs.~\eqref{eq:graphattention} and
\eqref{eq:graphmessage}.
Attention is restricted to selected neighbors and
self-connections, with zero edge contributions for
self-attention. Multiple heads are concatenated
before output projection and residual feed-forward
updates. Directed edge features are updated using
the input features of their two endpoints and their
previous edge representations
(Eq.~\eqref{eq:graphedge}).
Sparse connectivity restricts message propagation,
although attention is implemented using masked
dense tensors.

In WeaveVLA, Q-Former
uses learned queries to aggregate usable node features
and selected edge features into a fixed number of
relational tokens.
For StarVLA, relational tokens are appended
to the original VLM cross-attention context, preserving
its attention mask. For $\pi_{0.5}$, the projected
relational tokens are inserted before the noisy action
tokens, allowing the action expert to attend to both
relational information and the original vision-language
prefix.

\section{Training Policy}
\label{app:training_protocol}

We train WeaveVLA with parallel episode streams that preserve the temporal order of demonstration observations. Episode assignment and visit order are randomized. Let $b$ index a stream, $v$ a visit to its assigned demonstration, and $k$ the sampled-observation index within that visit. Using zero-based control-step indices, the initial offset and subsequent observation times are
\begin{equation}
\begin{aligned}
\delta_b^{(v)} &\leftarrow
\operatorname{Sample}\!\left(\{0,\ldots,\Delta-1\}\right),\\
t_{b,k}^{(v)} &= \delta_b^{(v)} + k\Delta,
\end{aligned}
\label{eq:training_temporal_sampling}
\end{equation}
where $\Delta$ is the training observation stride in control steps. The LIBERO and CALVIN configurations use $\Delta=8$. The sampling operator selects an offset at the start of each visit. The offset remains fixed throughout that visit, and only valid observation indices within the demonstration are used. Each batch contains one current observation from every active stream, paired with its demonstrated action chunk. The observation cursors advance after the batch is consumed.

Each stream maintains an independent graph state associated with its current episode and tracked instance identities. This state includes cached node features and pair-specific recurrent states. The stream-to-state correspondence is preserved across successive batches, so each observation receives history from its own demonstration. State values persist across optimization steps and are reused at the next sampled observation. Streams can reach episode boundaries at different times. When a stream starts a new demonstration visit, its graph state is reset before the first sampled observation, while the other streams retain their histories.

We truncate backpropagation at each sampled observation by treating the incoming graph state as a constant. Let $n$ index training batches, $\mathcal M_{b,n}^{\mathrm{in}}$ and $\mathcal M_{b,n}^{\mathrm{out}}$ denote the incoming and updated graph states, and $\mathcal F_{\theta_n}$ denote context construction and graph updating with the current model parameters. The current computation and state storage are
\begin{equation}
\begin{aligned}
\big(C_{b,n}^{\mathrm{act}},\mathcal M_{b,n}^{\mathrm{out}}\big)
&=\mathcal F_{\theta_n}\!\big(
o_{b,n},\ell_{b,n},
\operatorname{sg}(\mathcal M_{b,n}^{\mathrm{in}})\big),\\
\mathcal M_{b,n+1}^{\mathrm{in}}
&\leftarrow\operatorname{sg}(\mathcal M_{b,n}^{\mathrm{out}}),
\end{aligned}
\label{eq:training_state_propagation}
\end{equation}
where $o_{b,n}$ and $\ell_{b,n}$ are the observation and instruction assigned to stream $b$ in batch $n$, and $\operatorname{sg}$ denotes stop-gradient. The second line applies when the stream continues the same demonstration visit. The graph update follows the observation-validity, state-expiry, and reset rules of the model. The action expert uses $C_{b,n}^{\mathrm{act}}$, denoting the conditioning inputs for the selected backbone, to compute the current training loss. Gradients pass through the current graph computation, recurrent update, and action prediction, but stop at the incoming state. Detachment preserves the stored values while removing their computation history, giving a recurrent gradient window of one sampled update.

\smallskip
\noindent\textbf{Backbone-specific flow matching.}
The two action experts use different time parameterizations and Beta-derived training distributions. For StarVLA, the sampled noise-to-action time is
\begin{equation}
\tau_{\mathrm{G}}
=1-\frac{\min(z_{\mathrm{G}},s)}{s},
\qquad
z_{\mathrm{G}}\sim\operatorname{Beta}(1.5,1.0),
\quad s=0.999.
\label{eq:gr00t_time_sampling}
\end{equation}
This follows the noise-to-action convention in Sec.~\ref{sec:training}, with interpolation $A_t^\tau=(1-\tau)\epsilon+\tau A_t^*$ and velocity target $A_t^*-\epsilon$.

For $\pi_{0.5}$, the native action expert samples
\begin{equation}
t_{\pi}=0.999z_{\pi}+0.001,
\qquad
z_{\pi}\sim\operatorname{Beta}(1.5,1.0),
\label{eq:pi05_time_sampling}
\end{equation}
and constructs $A_t^{t_{\pi}}=t_{\pi}\epsilon+(1-t_{\pi})A_t^*$ with velocity target $\epsilon-A_t^*$. Under the common noise-to-action notation used in the main text, this corresponds to $\tau_{\pi}=1-t_{\pi}$ and the sign-reversed velocity field
\begin{equation}
v_\phi(\cdot,\tau_{\pi})
=-u_\phi(\cdot,1-\tau_{\pi}).
\label{eq:flow_time_conversion}
\end{equation}
This is a change of mathematical convention; each backbone retains its native time sampling and velocity prediction during implementation. The flow-matching loss supervises only action predictions, while relational tokens provide additional conditioning.

\section{Inference Algorithm}
\label{app:policyinference}

Algorithm~\ref{alg:graphvla_policy} summarizes policy updates at control steps $t$. Each update predicts $H$ actions and executes up to $n_{\mathrm{exec}}$ before replanning. Both values are set by the deployment configuration. For example, on LIBERO, $\pi_{0.5}$ predicts ten actions and executes the first eight before observing the scene again. CALVIN interrupts execution when the instruction changes. The set $\mathcal E_t^{\mathrm{obs}}$ contains edges with valid
current geometric support at both endpoints. Only these edges update
their recurrent states; other edges retain eligible history.
State validation applies the reset and expiry rules described above,
with zero initialization for new or invalidated pairs. CALVIN resets
graph states and task bindings at each new subtask.
$\operatorname{Cond}$ denotes edge-to-VLM conditioning.

The action sampler uses the interface described in Section~\ref{sec:conditioning}. In $\pi_{0.5}$, the graph uses the original vision-language prefix embeddings for conditioning. The attention mask allows the graph tokens to attend to the prefix and to one another. The action tokens attend to both blocks, while prefix attention remains unchanged. Graph construction and recurrent updating run once per observation. The resulting tokens are reused across denoising steps within the same action prediction.

\begin{center}
\begin{minipage}{0.85\linewidth}
\begin{algorithm}[H]
\caption{WeaveVLA inference with relational memory}
\label{alg:graphvla_policy}
\footnotesize
\setlength{\algorithmicindent}{1em}
\algrenewcommand\algorithmicrequire{\textbf{Input:}}
\begin{algorithmic}[1]
\Require Observations $o_t=(I_t,s_t)$, instruction $\ell_t$,
calibration $\mathcal C$, backbone $b$, $1\leq n_{\mathrm{exec}}\leq H$
\State Initialize persistent state $\mathcal S$
\For{each policy update $t$}
    \State Apply reset and expiry rules to $\mathcal S$
    \State $(H_t,F_t)\gets\operatorname{VLM}(I_t,\ell_t)$
    \State Build $\mathcal G_t$ and $\mathcal E_t^{\mathrm{obs}}$
    from $o_t,F_t,\ell_t,\mathcal C$
    \State $(\bar U_t,\bar E_t)\gets
    \operatorname{GNN}_{\mathrm{spatial}}(\mathcal G_t)$
    \For{each $(i,j)\in\mathcal E_t$}
        \State $m_{ij,t}\gets m_{ij,t^-}$
        \If{$(i,j)\in\mathcal E_t^{\mathrm{obs}}$}
            \State $m_{ij,t}\gets
            \operatorname{GRU}(\bar e_{ij,t},m_{ij,t^-})$
            \State Record the valid-update time $t$
        \EndIf
        \State $\tilde e_{ij,t}\gets
        \phi_M(\bar e_{ij,t}\mathbin{\Vert}m_{ij,t})$
    \EndFor
    \State $E_t^{\mathrm{cond}}\gets
    \operatorname{Cond}(\tilde E_t,H_t)$
    \State $(U_t,E_t)\gets
    \operatorname{GNN}_{\mathrm{task}}
    (\bar U_t,E_t^{\mathrm{cond}},\mathcal E_t)$
    \State $T_t^G\gets
    W_{\mathrm{out}}\operatorname{Readout}(Q,[U_t;E_t])$
    \State $A_t\gets\operatorname{SampleActions}_b(H_t,T_t^G,s_t,H)$
    \State Execute up to $n_{\mathrm{exec}}$ actions from $A_t$
\EndFor
\end{algorithmic}
\end{algorithm}
\end{minipage}
\end{center}

\section{Perception Adaptation}
\label{app:perception}

Instance masks are used to construct graph nodes and pool foreground
features. We obtain these masks by using YOLOE bounding boxes to prompt
a frozen SAM~3 model. To improve instance coverage within each benchmark,
we fine-tune separate YOLOE detectors for LIBERO and CALVIN D-to-D using
simulator-derived supervision. We evaluate both the detector outputs
and the resulting SAM~3 masks to assess how detector adaptation affects
the perception input to the graph module.

\begin{figure}[!hb]
\centering
\IfFileExists{figures/Figure3_paper_D.pdf}
  {\includegraphics[width=0.87\textwidth]{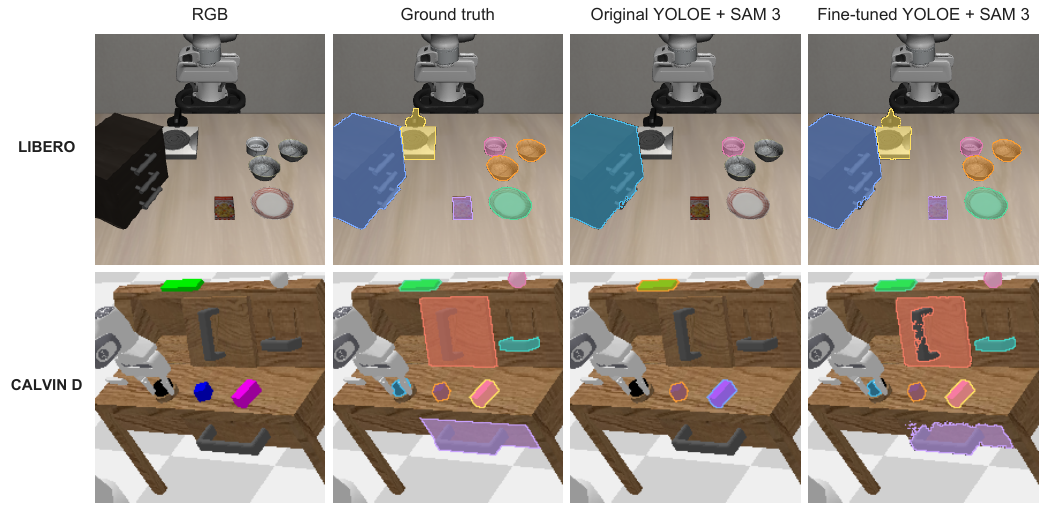}}
  {\includegraphics[width=0.87\textwidth]{Figure3_paper_D.pdf}}
\caption{Qualitative comparison of instance segmentation on LIBERO
(top) and CALVIN scene D (bottom). Columns show RGB observations,
simulator ground-truth masks, and masks produced by the same frozen
SAM~3 model using bounding boxes from the original and fine-tuned
YOLOE detectors. Fine-tuning recovers previously missed instances,
including the plate and cookie box in LIBERO and cabinet components
in CALVIN.}
\label{fig:qualitative}
\end{figure}

Simulator states provide paired static-camera RGB observations and
visible-instance masks for detector supervision. For LIBERO, the
dataset is constructed from 2,000 demonstrations covering 28 canonical
object categories. Before splitting, we group demonstrations by hashes
of their complete state and action sequences to keep identical recorded
trajectories within the same partition. We remove robot pixels from
the detector inputs and convert visible-instance masks into polygon
annotations. A frame is discarded if any retained object's polygon
has an IoU below 0.90 with its source mask. This filtering removes
182 frames and yields 12,599 training, 1,578 validation, and 1,641 test
frames. The test split contains held-out demonstrations with task
categories and simulator assets shared with the training split.

For CALVIN, we fine-tune the detector directly on scene-D demonstrations
using 11 categories covering scene objects and robot parts. We split
demonstration episodes before sampling frames and render the recorded
robot and scene states to obtain RGB observations and instance
annotations. The dataset contains 256 training frames from 24 episodes
and 64 validation frames from seven disjoint episodes. All 320 RGB
images are unique, and exact RGB matches to the 40 scene-D comparison
frames used below are excluded. Polygon annotations retain contours
with at least three vertices and an area of at least two pixels.

Both adaptations start from the original YOLOE-11L segmentation
checkpoint and use AdamW with an input resolution of 640. For LIBERO,
we use an initial learning rate of $2\times10^{-4}$, a batch size of
16, two warm-up epochs, and up to 30 training epochs, with an
early-stopping patience of eight epochs. For CALVIN, we use an initial
learning rate of $1.5\times10^{-4}$, a batch size of four, one warm-up
epoch, and eight training epochs. We select the checkpoint with the
best validation performance for each benchmark.

We evaluate the box and mask predictions produced by YOLOE itself
in Table~\ref{tab:perception}. Within each benchmark, the original
and fine-tuned detectors use the same evaluation images and canonical
vocabulary. LIBERO results are computed on the 1,641 held-out test
frames retained after polygon-quality filtering, using a confidence
threshold of 0.001 and a non-maximum-suppression IoU threshold of 0.5.
CALVIN results are computed on the same 64 scene-D validation frames
used for checkpoint selection, with a configured
non-maximum-suppression IoU threshold of 0.7. Fine-tuning improves
all reported box and mask AP metrics under these evaluation protocols.

\begin{table}[!ht]
\centering
\caption{Box and mask AP (\%) of YOLOE. LIBERO uses 1,641 held-out
test frames. CALVIN compares both checkpoints on the same 64 scene-D
validation frames used for checkpoint selection.}
\label{tab:perception}
{\papertablesetup
\begin{tabular*}{0.82\linewidth}{@{\extracolsep{\fill}}l|c|c|c|c@{}}
\toprule
 & \multicolumn{2}{c|}{\tablehead{Box AP $\uparrow$}}
 & \multicolumn{2}{c}{\tablehead{Mask AP $\uparrow$}}\\
\tablehead{Method}
 & \tablehead{50}
 & \tablehead{50--95}
 & \tablehead{50}
 & \tablehead{50--95}\\
\midrule
\multicolumn{5}{c}{\tablehead{LIBERO}}\\
\midrule
Original YOLOE
 & 25.8 & 23.5 & 25.8 & 20.2\\
Adapted YOLOE
 & \textbf{99.5}
 & \textbf{99.0}
 & \textbf{99.5}
 & \textbf{93.4}\\[-2pt]
 & {\scriptsize\textcolor[rgb]{0.12,0.45,0.36}{(+73.7)}}
 & {\scriptsize\textcolor[rgb]{0.12,0.45,0.36}{(+75.5)}}
 & {\scriptsize\textcolor[rgb]{0.12,0.45,0.36}{(+73.7)}}
 & {\scriptsize\textcolor[rgb]{0.12,0.45,0.36}{(+73.2)}}\\[2pt]
\midrule
\multicolumn{5}{c}{\tablehead{CALVIN}}\\
\midrule
Original YOLOE
 & 29.8 & 23.9 & 29.8 & 21.4\\
Adapted YOLOE
 & \textbf{96.7}
 & \textbf{91.3}
 & \textbf{96.6}
 & \textbf{85.9}\\[-2pt]
 & {\scriptsize\textcolor[rgb]{0.12,0.45,0.36}{(+66.9)}}
 & {\scriptsize\textcolor[rgb]{0.12,0.45,0.36}{(+67.4)}}
 & {\scriptsize\textcolor[rgb]{0.12,0.45,0.36}{(+66.8)}}
 & {\scriptsize\textcolor[rgb]{0.12,0.45,0.36}{(+64.5)}}\\[2pt]
\bottomrule
\end{tabular*}}
\end{table}

To evaluate the masks supplied to graph construction, we compare the
original and fine-tuned detectors with SAM~3 held fixed. The comparison
uses one held-out test frame from each of the 40 LIBERO tasks and
40 CALVIN scene-D frames rendered from the initial states of benchmark
evaluation sequences. Each RGB observation is paired with simulator
instance annotations.

Both pipelines use a detector confidence threshold of 0.30, a
non-maximum-suppression IoU threshold of 0.5, a SAM~3 quality threshold
of 0.25, and a minimum visible-mask area of 48 pixels. After removing
robot pixels, predicted masks are matched one-to-one to visible
ground-truth instances of the same category based on mask overlap.
We average IoU over ground-truth instances, assigning zero to missed
instances and excluding unmatched predictions from the average.
Evaluation images, category mappings, SAM~3 weights, and
post-processing thresholds are identical between the two pipelines
within each benchmark.

\begin{table}[!ht]
\centering
\caption{Mean object-mask IoU (\%) with a fixed SAM~3 segmenter on
40 matched frames per benchmark. Missed objects receive zero IoU.}
\label{tab:sam}
{\papertablesetup
\begin{tabular*}{0.5\linewidth}{@{\extracolsep{\fill}}lrr@{}}
\toprule
\tablehead{Benchmark}
 & \tablehead{Original}
 & \tablehead{Adapted}\\
\midrule
LIBERO
 & 12.1
 & \textbf{97.0}\enspace{\footnotesize\textcolor[rgb]{0.12,0.45,0.36}{(+84.9)}}\\
\addlinespace[2pt]
CALVIN D
 & 18.6
 & \textbf{91.1}\enspace{\footnotesize\textcolor[rgb]{0.12,0.45,0.36}{(+72.4)}}\\
\bottomrule
\end{tabular*}}
\end{table}

As shown in Table~\ref{tab:sam}, fine-tuning increases mean
object-mask IoU from 12.1\% to 97.0\% on LIBERO and from 18.6\%
to 91.1\% on CALVIN scene D. Figure~\ref{fig:qualitative} further
shows that the fine-tuned detector identifies all annotated
objects in the illustrated scenes. The adapted detector provides
bounding-box prompts for previously missed objects, and the
frozen SAM~3 model uses these prompts to generate their instance
masks.

\section{Policy Ablation Protocols}
\label{app:policyablations}

Table~\ref{tab:policy} compares the original StarVLA-GR00T backbone, full WeaveVLA, and its three ablated variants on LIBERO and LIBERO-Plus. 
The three ablated WeaveVLA variants use the same backbone, training data, optimization budget, and checkpoint-selection rule as the full model. 
Evaluation uses matched initial states and 500 rollouts per LIBERO suite, while LIBERO-Plus reports pooled success rates across perturbations without additional fine-tuning. 
Across the full model and its three ablations, perception, tracked identities, TCP anchoring, coverage-weighted appearance pooling, and the original VLM context are kept consistent.

Removing geometry reconstruction sets $\lambda_g=0$, retaining the flow-matching objective and all geometric inputs. The no-edge-memory variant removes the GRU and its pair-specific hidden states. Spatial edge features are passed to the subsequent VLM-conditioning and readout stages without recurrent fusion, while geometry caching and graph connectivity remain unchanged.

The full-connectivity variant replaces task-conditioned connections with directed edges between all distinct valid scene nodes, including task-irrelevant objects and the gripper. For $N$ valid nodes, this produces $N(N-1)$ directed edges without self-loops. No additional ground-truth objects are introduced. Relative geometric descriptors, recurrent edge updates, observation-validity rules, and the remaining model components are preserved.

\section{Real-World Evaluation Details}
\label{app:realworld-progress}

\noindent\textbf{Evaluation protocol.}
The protocol specifies 30 standard trials and 30 perturbed trials per task and method. The perturbed trials allocate ten trials each to table clutter, random lighting, and background variation.

\smallskip
\noindent\textbf{Task instructions.}\par
\noindent\textbf{Task 1.} ``Stack the bowls on the plate.''\par
\noindent\textbf{Task 2.} ``Place the pens in the cups.''\par
\noindent\textbf{Task 3.} ``Hang the mug on the rack, then put the first crumpled paper ball, the block, and the second crumpled paper ball into the basket, in that order.''\par

\smallskip
\noindent\textbf{Progress scoring.}
The progress score starts at zero and counts consecutively completed milestones in the order specified in Table~\ref{tab:realworld-progress}. Binary task success is reported separately and requires completion of all milestones (3, 4, and 4 for Tasks 1--3, respectively).

\begin{table}[!htbp]
\centering
\captionsetup{hypcap=false}

\caption{Real-world progress scoring rubric. Each fraction gives the milestone score over the maximum task score.}
\label{tab:realworld-progress}

{\papertablesetup
\footnotesize
\setlength{\tabcolsep}{5pt}
\renewcommand{\arraystretch}{1.08}
\begin{tabular}{@{}p{.22\columnwidth}p{\dimexpr.78\columnwidth-2\tabcolsep\relax}@{}}
\toprule
\tablehead{Task} & \tablehead{Ordered milestones} \\
\midrule

\raggedright Bowl Stacking
& \raggedright
  \textbf{(1/3)} Grasp the first bowl and place it on the plate.
  \textbf{(2/3)} Grasp the second bowl and align it with the first bowl's actual position.
  \textbf{(3/3)} Stack it stably on the first bowl, release it, and withdraw the gripper.
  \tabularnewline

\addlinespace[2pt]

\raggedright Place Pens in Cups
& \raggedright
  \textbf{(1/4)} Grasp the first pen and lift it clear of the table.
  \textbf{(2/4)} Place it in its target cup and release it.
  \textbf{(3/4)} Grasp the second pen and lift it clear of the table.
  \textbf{(4/4)} Place it in its target cup and release it, with both pens correctly placed.
  \tabularnewline

\addlinespace[2pt]

\raggedright Clean Table
& \raggedright
  \textbf{(1/4)} Hang the mug securely on the rack.
  \textbf{(2/4)} Put the first crumpled paper ball in the basket.
  \textbf{(3/4)} Put the block in the basket.
  \textbf{(4/4)} Put the second crumpled paper ball in the basket.
  \tabularnewline

\bottomrule
\end{tabular}
}

\end{table}

\begin{table}[!htbp]
\centering
\caption{Real-world success rates (\%) by perturbation. Each task contributes ten trials per method and perturbation. The last block pools the three tasks. StarVLA denotes StarVLA-GR00T.}
\label{tab:realworld-ood}

{\papertablesetup
\footnotesize
\setlength{\tabcolsep}{4pt}
\renewcommand{\arraystretch}{1.05}
\begin{tabular*}{\columnwidth}{@{\extracolsep{\fill}}llrrr@{}}
\toprule
\tablehead{Task} & \tablehead{Method}
& \tablehead{Clutter}
& \tablehead{Lighting}
& \tablehead{Background} \\
\midrule

Bowl stacking & WeaveVLA & 80.0 & 70.0 & 80.0 \\
 & $\pi_{0.5}$ & 60.0 & 60.0 & 70.0 \\
 & StarVLA & 60.0 & 50.0 & 50.0 \\

\addlinespace[2pt]

Pen placement & WeaveVLA & 30.0 & 40.0 & 70.0 \\
 & $\pi_{0.5}$ & 40.0 & 20.0 & 40.0 \\
 & StarVLA & 20.0 & 40.0 & 30.0 \\

\addlinespace[2pt]

Table cleaning & WeaveVLA & 30.0 & 40.0 & 40.0 \\
 & $\pi_{0.5}$ & 30.0 & 20.0 & 30.0 \\
 & StarVLA & 0.0 & 30.0 & 20.0 \\

\midrule

All tasks & WeaveVLA
& \textbf{46.7}
& \textbf{50.0}
& \textbf{63.3} \\
 & $\pi_{0.5}$ & 43.3 & 33.3 & 46.7 \\
 & StarVLA & 26.7 & 40.0 & 33.3 \\

\bottomrule
\end{tabular*}
}

\end{table}

\FloatBarrier

\section{Spatial Relations During Manipulation}
\label{app:spatialvisualization}
\noindent\begin{minipage}{\linewidth}
\centering
\includegraphics[width=.30\linewidth,trim=18bp 464bp 892bp 69bp,clip]{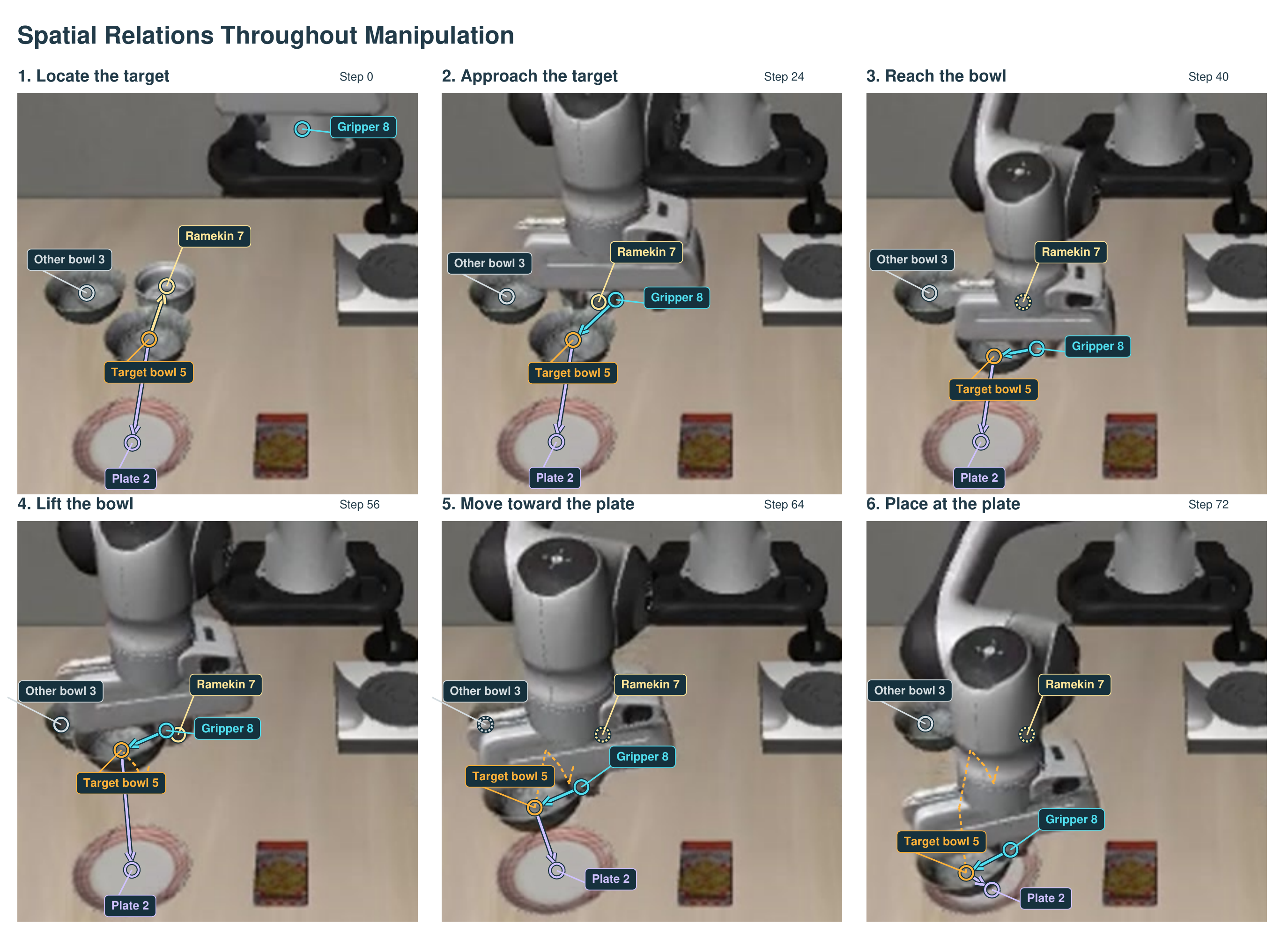}\hspace{.025\linewidth}
\includegraphics[width=.30\linewidth,trim=453bp 464bp 457bp 69bp,clip]{figures/spatial_relations.pdf}\hspace{.025\linewidth}
\includegraphics[width=.30\linewidth,trim=888bp 464bp 22bp 69bp,clip]{figures/spatial_relations.pdf}\par
\includegraphics[width=.30\linewidth,trim=18bp 26bp 892bp 507bp,clip]{figures/spatial_relations.pdf}\hspace{.025\linewidth}
\includegraphics[width=.30\linewidth,trim=453bp 26bp 457bp 507bp,clip]{figures/spatial_relations.pdf}\hspace{.025\linewidth}
\includegraphics[width=.30\linewidth,trim=888bp 26bp 22bp 507bp,clip]{figures/spatial_relations.pdf}
\captionsetup{hypcap=false}\captionof{figure}{Spatial relations throughout a crowded manipulation task. Six observations from a successful WeaveVLA rollout track the same gripper, target bowl, plate, and nearby objects. Arrows show selected recorded graph connections, and the dashed orange trace follows the target bowl in the image plane.}
\label{fig:spatialvisualization}
\end{minipage}
\par\smallskip
\setlength{\baselineskip}{10.5pt}{Figure~\ref{fig:spatialvisualization} shows a LIBERO-Spatial rollout (task 0, trial 24) in which the robot picks up the black bowl between the plate and the ramekin and places it on the plate. The six panels correspond to control steps 0, 24, 40, 56, 64, and 72. Persistent entity IDs link the target bowl (5), neighboring bowl (3), plate (2), ramekin (7), and gripper (8) across observations. Both black bowls are instruction candidates in the recorded graph.

Initially, the target bowl lies between the plate and ramekin in the scene. As the gripper approaches and lifts it, the gripper--bowl relation changes along with the bowl's position relative to the references. The final panels show the bowl moving toward and overlapping the plate in the image. The displayed arrows are selected directed connections from the graph logs. Colors and the choice of highlighted edges serve illustration. Dashed circles denote cached node locations during missing observations. The overlays use logged image coordinates and depict image-plane spatial relationships.
\par}

\end{document}